\documentclass[runningheads]{llncs}
\usepackage{comment}
\usepackage{times}
\usepackage{url}
\usepackage{latexsym}

\usepackage{graphicx}  
\usepackage{bm}
\usepackage{amsmath, amssymb, amsfonts}
\usepackage{multirow}
\usepackage{multicol}
\usepackage{enumitem}

\usepackage{floatrow}
\newfloatcommand{capbtabbox}{table}[][\FBwidth]
\begin{document}
\title{DGA-Net: Dynamic Gaussian Attention Network for \\Sentence Semantic Matching}

\def\YOFOSubNumber{11}  

\newcommand{\shortname}{\emph{DGA-Net}}
\newcommand{\oname}{\emph{Dynamic Gaussian Attention Network}}
\newcommand{\fullname}{\emph{Dynamic Gaussian Attention Network (DGA-Net)}}

\titlerunning{DGA-Net for Sentence Semantic Matching}
%
\author{Kun Zhang\inst{1,2}\thanks{Corresponding Author} \and
Guangyi Lv\inst{3} \and
Meng Wang\inst{1,2} \and
Enhong Chen\inst{3}}
\authorrunning{Kun et al.}
%
\institute{Key Laboratory of Knowledge Engineering with Big Data, Hefei University of Technology, China \and
School of Computer Science and Information Engineering, Hefei University of Technology, China \and
School of Computer Science and Technology, University of Science and Technology of China, China
\\
\email{\{zhang1028kun, eric.mengwang\}@gmail.com}\\
\email{\{gylv, cheneh\}@mail.ustc.edu.cn}}
\maketitle              

\begin{abstract}
	Sentence semantic matching requires an agent to determine the semantic relation between two sentences, where much recent progress has been made by advancement of representation learning techniques and inspiration of human behaviors. 
	Among all these methods, attention mechanism plays an essential role by selecting important parts effectively. 
	However, current attention methods either focus on all the important parts in a static way or only select one important part at one attention step dynamically, which leaves a large space for further improvement. 
	To this end, in this paper, we design a novel \fullname~to combine the advantages of current static and dynamic attention methods. 
	More specifically, we first leverage pre-trained language model to encode the input sentences and construct semantic representations from a global perspective. 
	Then, we develop a Dynamic Gaussian Attention~(DGA) to dynamically capture the important parts and corresponding local contexts from a detailed perspective. 
	Finally, we combine the global information and detailed local information together to decide the semantic relation of sentences comprehensively and precisely. 
	Extensive experiments on two popular sentence semantic matching tasks demonstrate that our proposed \shortname~is effective in improving the ability of attention mechanism. 
\end{abstract}

\section{Introduction}
\label{s:introduction}
\begin{figure}
	\centering
	\includegraphics[width=0.6\textwidth]{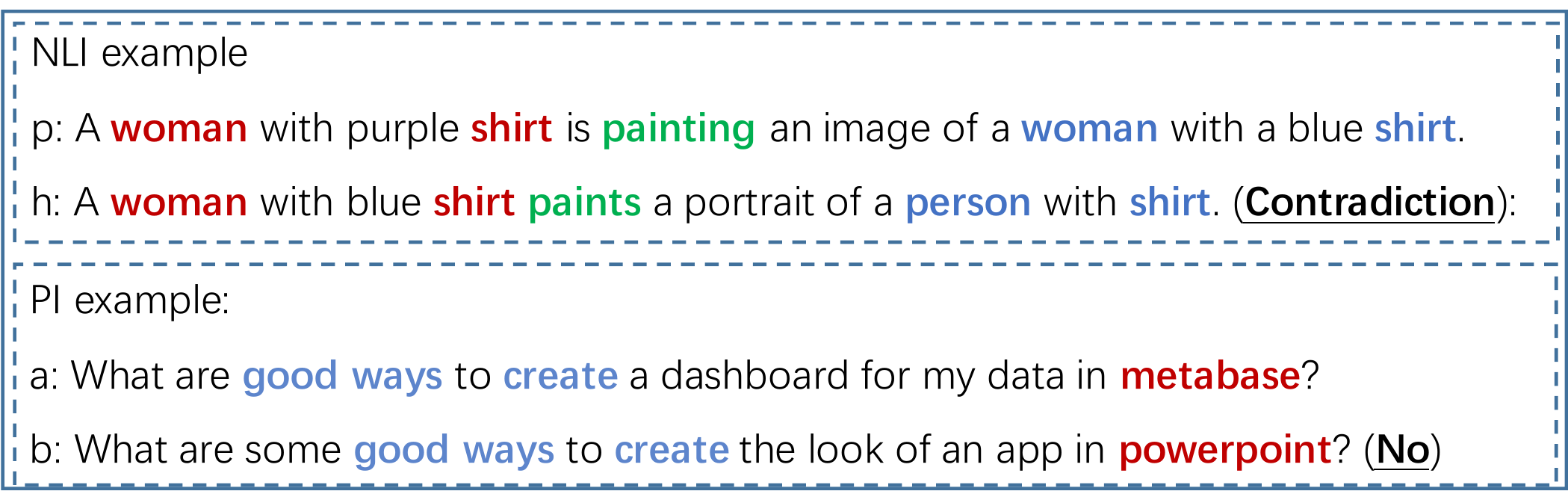}
	\caption{Two example from different sentence semantic matching datasets (colored words are the important parts that need attention).}
	\label{f:example}
\end{figure}

Sentence semantic matching is a long-lasting theme of Natural Language Processing~(NLP), which requires an agent to determine the semantic relations between two sentences. 
For example, in Natural Language Inference~(NLI), it is used to determine whether a hypothesis  can be inferred reasonably from a given premise~\cite{Kim2018SemanticSM}. 
In Paraphrase Identification~(PI), it is utilized to identify whether two sentences express the same meaning or not~\cite{dolan2005automatically}.  
Fig.~\ref{f:example} gives us two representative examples of NLI and PI.

As a fundamental technology, sentence semantic matching has been applied successfully in many NLP fields,  e.g., information retrieval~\cite{Clark2016CombiningRS,sun2020dual}, question answering~\cite{liu2018finding}, and dialog system~\cite{serban2016building}. 
With advanced representation learning techniques~\cite{devlin2018bert,hong2019deep,wu2020joint}, numerous efforts have been dedicated to this task, where the dominant trend is to 
build complex structures with attention. 
For example, self-attention~\cite{vaswani2017attention} can generate better representations by relating elements at different positions in a single sentence. 
Co-attention~\cite{Kim2018SemanticSM,zhang2019context} focuses on sentence interaction from a detailed perspective.
Dynamic re-read attention~\cite{zhang2019drr} is able to select the important parts in a dynamic way based on learned information. 
They all help to achieve impressive performance.

However, most work either focuses on all the important parts in a static way~\cite{Chen-Qian2017ACL} or only selects one important part at each selection in a dynamic way~\cite{zhang2019drr}. 
They either are incapable of adapting to dynamic changes during the sentence understanding process or ignore the importance of local structures. 
For example, in Fig.~\ref{f:example}, colored words illustrate the focus points. 
When selecting the important parts as the static attention methods do, the representations of two sentences may be similar since many of the important words are the same (e.g.,\textit{woman, shirt}). 
When employing the dynamic attention methods~\cite{zhang2019drr}, the attributes of the selected parts may be missed since dynamic methods only select one important word at each step and ignore the local contexts (e.g.,\textit{woman with purple shirt, woman with blue shirt}). 
All these will lead to a wrong decision. 
Therefore, how to leverage attention mechanism to select proper information for precise sentence semantic understanding and matching is the main challenge that we need to consider.

To this end, in this paper, we propose an effective \fullname~approach to combine the advantages of current static and dynamic attention methods.
In concerned details, we first utilize pre-trained BERT to model the semantic meanings of input words and sentences globally.  
Based on the dynamic attention mechanism and Gaussian distribution, we develop a novel Dynamic Gaussian Attention~(DGA) to pay close attention to one important part and corresponding local contexts among sentences at each attention step simultaneously.  
Along this line, we can not only focus on the most important part of sentences dynamically, but also use the local context to support the understanding of these selected parts precisely. 
Extensive evaluations on two popular sentence semantic matching tasks (i.e., NLI and PI) demonstrate the effectiveness of our proposed \shortname~method and its advantages over state-of-the-art sentence encoding-based baselines.

\section{Related Work}
\label{s:related-work}

With the available large annotated datasets, such as SNLI~\cite{bowman2015large}, SCITAIL~\cite{khot2018scitail}, and Quora Question Pair~\cite{iyer2017first}, as well as various neural networks, such as LSTM~\cite{Cheng2016LongSM}, GRU~\cite{Chung2014EmpiricalEO}, and attention mechanism~\cite{Parikh2016ADA,vaswani2017attention,zhang2019drr,zhang2019multilevel,zhang2021making}, plenty of methods have been developed to represent and evaluate sentence semantic meanings.  
Among all methods, attention mechanism has become the essential module, which helps models capture semantic relations and properly align the elements of sentences. 
For example, Liu et al.~\cite{Liu2016LearningNL} proposed inner-attention to pay more attention to the important words among sentences. 
In order to better capture the interaction of sentences, Kim et al.~\cite{Kim2018SemanticSM} utilized co-attention network to model the interaction among sentence pairs.
Moreover, Cho~\cite{im2017distance} and Shen~\cite{shen2017disan} proposed to utilize multi-head attention to model sentence semantics and interactions from multiple aspects without RNN/CNN structure. 
They took full advantage of attention mechanism for better sentence semantic modeling and achieved impressive performance on sentence semantic matching task. 

Despite the success of using attention mechanism in a static way, researchers also learn from human behaviors and propose dynamic attention methods.
By conducting a lab study, Zheng et al.~\cite{zheng2019human} observed that users generally read the document from top to bottom with the reading attention decays monotonically. 
Moreover, in a specific scenario~(e.g., Answer Selection), users tend to pay more attention to the possible segments that are relevant to what they want. They will reread more snippets of candidate answers with more \textit{skip} and \textit{up} transition behaviors, and ignore the irrelevant parts~\cite{li2019teach}. 
Furthermore, Zhang et al.~\cite{zhang2019drr} designed a novel dynamic re-read attention to further improve model performance. They tried to select one important word at each attention calculation and repeated this operation for precise sentence semantic understanding. 

However, static attention methods select all the important parts at one time, which may lead to a misunderstanding of sentence semantics since there are too many similar but semantically different important parts. 
Dynamic methods only select one important part at each operation, which may lose some important attributes of the important parts. 
Thus, we propose a \shortname~to select the important parts and corresponding local context in sentences for better sentence semantic understanding and matching.

\section{Problem Statement and Model Structure}
\label{s:model}

In this section, we formulate the NLI task as a supervised classification problem and introduce the structure and technical details of our proposed \shortname. 

\subsection{Problem Statement}
\label{s:problem-statement}
First, we define our task in a formal way. 
Given two sentences $\bm{s}^a = \{\bm{w}_1^a, \bm{w}_2^a, ..., \bm{w}_{l_a}^a \}$ and $\bm{s}^b = \{\bm{w}_1^b, \bm{w}_2^b, ..., \bm{w}_{l_b}^b \}$. Our goal is to learn a classifier $\xi$ which is able to precisely predict the relation $y = \xi(\bm{s}^a, \bm{s}^b)$ between $\bm{s}^a$ and $\bm{s}^b$.
Here, $\bm{w}_i^a$ and $\bm{w}_j^b$ are one-hot vectors which represent the $i^{th}$ and $j^{th}$ word in the sentences. 
$l_a$ and $l_b$ indicate the total number of words in $\bm{s}^a$ and $\bm{s}^b$, respectively.

\begin{figure*}
	\centering
	\includegraphics[width=0.95\textwidth]{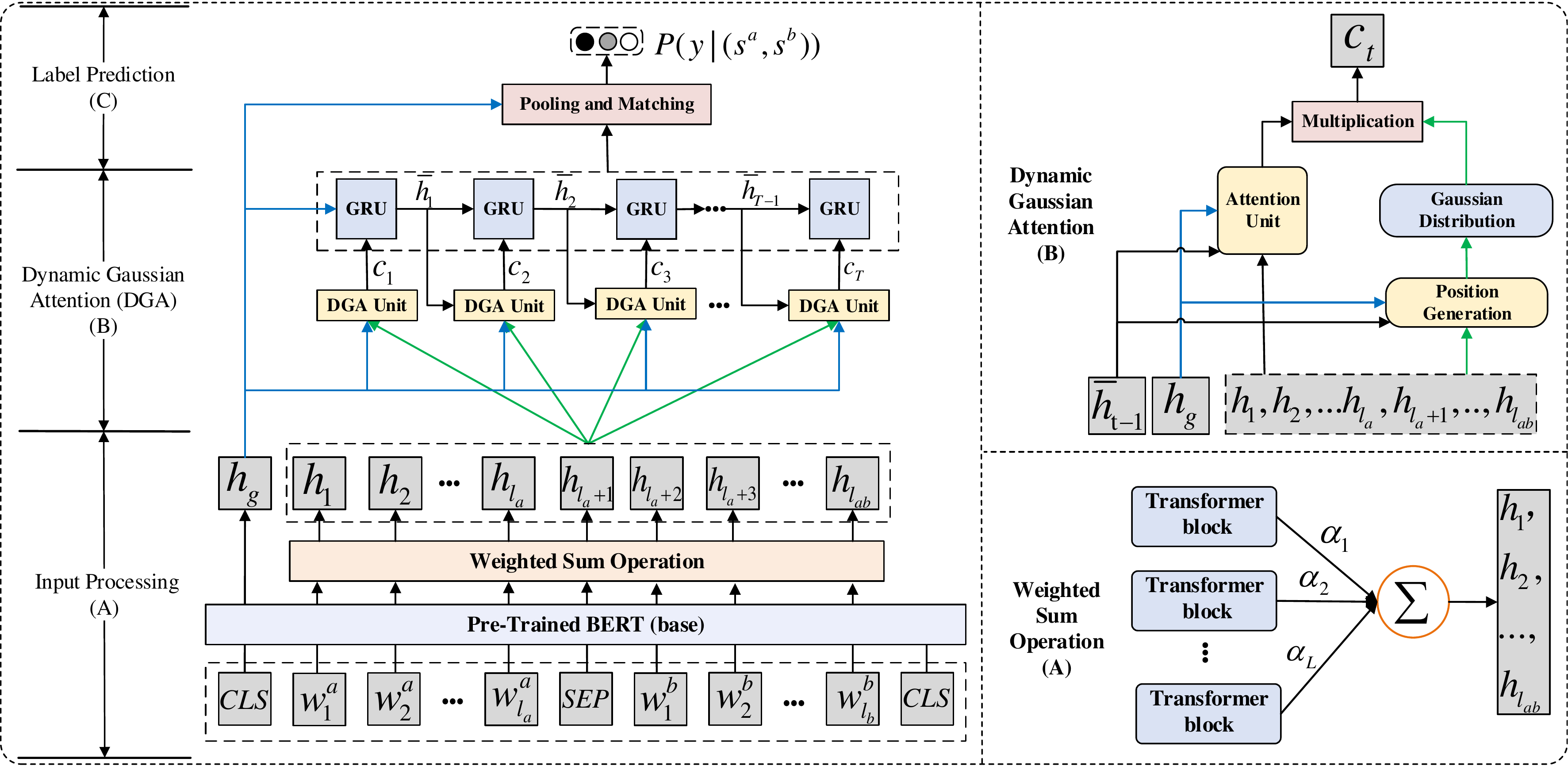}
	\caption{Architecture of \fullname.} 
	\label{f:model}
\end{figure*}

In order to model sentence semantic meanings more precisely and comprehensively, the following important challenge should be considered:
\begin{itemize}\setlength{\itemsep}{0pt}
	\item How to overcome the shortcomings of static and dynamic attention methods, and leverage attention operation to select proper information for precise sentence semantic understanding and matching?
\end{itemize}

To this end, we propose a novel \fullname~to tackle the above issue and doing better sentence semantic matching.

\subsection{\oname}
The overall architecture of \shortname~is shown in Figure~\ref{f:model}, which consists of three main components: 
1) \textit{Input Processing}: utilizing pre-train BERT to generate the extravagant representation of input words; 
2) \textit{Dynamic Gaussian Attention}: selecting one important part and proper local structure at each step and dynamically reading these contextual parts with all learned information;
3) \textit{Label Prediction}: predicting the final results based on the expanded semantic representations.

\textbf{Input Processing}
\label{s:input}
By making full use of large corpus and multi-layer transformers, BERT~\cite{devlin2018bert} has accomplished much progress in many natural language tasks and become a powerful tool to process the raw input sentences. 
Therefore, we also employ BERT to encode the input sentences.  
In order to make full use of BERT and encode sentence comprehensively, we use the weighted sum of all the hidden states from different transformer layers of BERT as the final contextual representations of input sentences. 
Specifically, the input sentence $\bm{s}^a = \{\bm{w}_1^a, \bm{w}_2^a, ..., \bm{w}_{l_a}^a \}$ and $\bm{s}^b = \{\bm{w}_1^b, \bm{w}_2^b, ..., \bm{w}_{l_b}^b \}$ will be split into BPE tokens~\cite{sennrich2015neural}. 
Then, we leverage a special token \textit{``[SEP]''} to concatenate two sentences and add \textit{``[CLS]''} token at the beginning and the end of concatenated sentences. 
As illustrated in Fig.~\ref{f:model}(A), suppose the final number of tokens in the sentence pair is $l_{ab}$, and BERT generates $L$ hidden states for each BPE token $\bm{BERT}_t^l, (1 \leq l \leq L, 1 \leq t \leq l_{ab})$. 
The contextual representation for $t^{th}$ token in input sentence pair at token level is then a per-layer weighted sum of transformer block output, with weights $\alpha_1, \alpha_2,...,\alpha_L$.
\begin{equation}
	\label{eq:input-processing}
	\begin{split}
		\bm{h}_t &= \sum_{l=1}^{L}\alpha_l\bm{\bm{BERT}_t^l}, \quad 1 \leq t \leq l_{ab}, \\ 
	\end{split}
\end{equation} 
where $\alpha_l$ is the weight for the $l^{th}$ layer in BERT and will be learned during the training. 
$\bm{h}_t$ is the representation for the $t^{th}$ token.
Moreover, we treat the output $\bm{BERT}_0^L$ of the first special token \textit{``[CLS]''} in the last block as the contextual representation $\bm{h}_g$ for input sentences globally. 
Along this line, we can model the semantic meanings of words and sentences comprehensively, which lays a good foundation for subsequent study.

\begin{figure}
	\centering
	\includegraphics[width=0.92\textwidth]{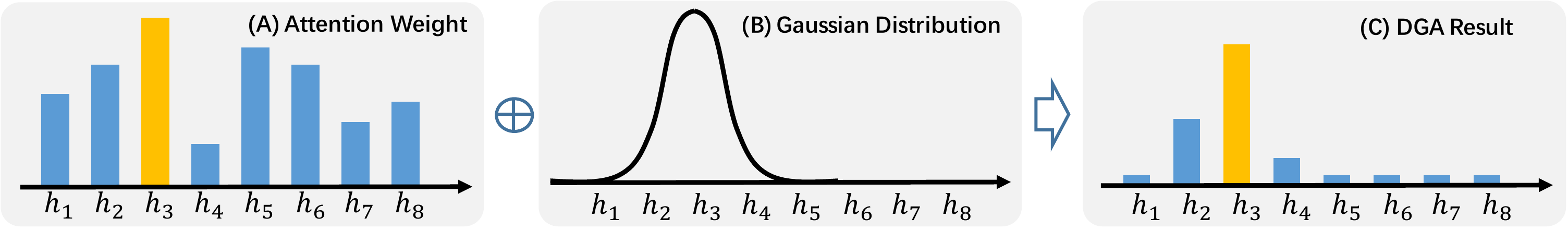}
	\caption{The processing of Dynamic Gaussian Attention~(DGA) calculation.}
	\label{f:dga_sim}
\end{figure}

\textbf{Dynamic Gaussian Attention}
\label{s:local-attention}
As introduced in Section~\ref{s:introduction}, static attention methods select all the important parts at one time, which may lead to a misunderstanding of sentence semantic meanings since there are too many similar but semantically different important parts. 
Meanwhile, dynamic attention methods~\cite{zhang2019drr} try to select one important part at step, which can alleviate the problem that static attention methods suffer from.  
However, it still causes the model to lose some important attributes of important parts and lead to an incorrect result. 
Therefore, it is crucial to employ attention mechanism in a proper way for better sentence semantic understanding and matching.

Inspired by previous work~\cite{zhang2019drr,zheng2019human} and Gaussian distribution, we design a novel \textit{Dynamic Gaussian Attention~(DGA)} unit to select the important part and proper local context simultaneously. 
Fig.~\ref{f:model}(B) and Fig.~\ref{f:dga_sim} illustrate the entire processing of DGA calculation. 
During each DGA operation, we first calculate the attention weight among input sequence. 
Meanwhile, we leverage a position generation method $\mathrm{G}(\cdot)$ to predict the focus point, which can be visualized as the yellow bar in Fig.~\ref{f:dga_sim}. 
Then, we generate a Gaussian distribution with the focus point as the center. 
Next, we multiply the attention weight and Gaussian distribution to get the DGA result. 
Along this line, the attention weights of the words that are close to the important part will be preserved, and the rest will be discarded. 
In other words, we can focus on the important part and corresponding local context for better semantic understanding. 
Inspired by DRr-Net~\cite{zhang2019drr}, we also repeat DGA operation many times for the final decision.

Specifically, DGA unit treats  $\{\bm{h}_i|i=1, 2, ..., l_{ab}\}$ as the inputs, and produces an important position $p_t$ at attention step $t$. 
The representation $\bm{c}_t$ for this position is derived as a weighted summation over the inputs within the window $[p_t-\frac{D}{2}, p_t+\frac{D}{2}]$; $D$ is the window size.  
Since we select these important parts in a sequential manner, GRU is adopted to encoder these important parts. 
This process can be formulated as follows:
\begin{equation}
	\label{eq:overall-local}
	\begin{split}
		\bm{H} &= [\bm{h}_1, \bm{h}_2, ..., \bm{h}_{l_{ab}}], \quad
		p_t = \mathrm{G}(\bm{H}, \bar{\bm{h}}_{t-1}, \bm{h}_g), \\
		\bm{c}_t &= \mathrm{F}(p_t, \bm{H}, \bar{\bm{h}}_{t-1}, \bm{h}_g), \quad
		\bar{\bm{h}}_t = \mathrm{GRU}(\bar{\bm{h}}_{t-1}, \bm{c}_t), \quad t = 1, 2, ..., T, 
	\end{split}
\end{equation}
where $\mathrm{G}(\cdot)$ is position generation function. 
$\mathrm{F}(\cdot)$ denotes DGA function. 
$T$ is the dynamic attention length. 
In order to take global information into consideration, we also treat the global representation $\bm{h}_g$ as an additional context in $\mathrm{G}(\cdot)$ and $\mathrm{F}(\cdot)$. 
$\bar{\bm{h}}_T$ can be regarded as the dynamic locally-aware representation for the input sentence pair.

Different from DRr-Net~\cite{zhang2019drr} that treats the word that has biggest weight as the current selection, we intend to use MLP to predict the focus point at current step.  
More specifically, we utilize position generation function $\mathrm{G}(\cdot)$ to generate the important position $p_t$ at attention step $t$ as follows:
\begin{equation}
	\label{eq:position-generation}
	\begin{split}
		\bm{m}_t &= \sum_{i=1}^{l_{ab}}(\bm{W}_1^p\bm{h}_i) + \bm{W}_2^p\bar{\bm{h}}_{t-1} + \bm{W}_3^p\bm{h}_g, \\
		p_t &= l_{ab} \cdot \mathrm{sigmoid}(\bm{v}_p^{\mathrm{T}}\mathrm{tanh}(\bm{U}_p\bm{m}_t)),
	\end{split}
\end{equation}
where $\{\bm{W}_1^p, \bm{W}_2^p, \bm{W}_3^p, \bm{v}_p, \bm{U}_p\}$ are trainable parameters. $\mathrm{T}$ is transposition operation. 
As the result of $\mathrm{sigmoid}(\cdot)$ function, $p_t \in [0, l_{ab}]$. 
Along this line, we are able to use all the learned information to generate the important position at each attention step. 

After getting the important position $p_t$, it is urgent to ensure its exact meaning in the sentence, which is in favor of overcoming the issue in section~\ref{s:problem-statement}.
Inspired by the observation that adjacent words contribute more for understanding current phrase than distant ones, we develop a novel DGA method by placing a Gaussian distribution centered around $p_t$ to further process the attention weights. 
The implementation function $\mathrm{F}(\cdot)$ can be formulated as follows:
\begin{equation}
	\label{eq:local-attention}
	\begin{split}
		\bm{g}_t=  &exp(-\frac{(\bm{s} - p_t)^2}{2\sigma^2}), \\
		\bm{\alpha}^a = &\omega_d^T\mathrm{tanh}(\bm{W}_d\bm{H} + (\bm{U}_d\bar{\bm{h}}_{t-1}+ \bm{M}_d\bm{h}_g)\otimes \bm{e}_{l_{ab}}), \\
		\bar{\bm{\alpha}}^a = &\bm{\alpha}^a \cdot \bm{g}_t, \quad
		\bm{c}_t = \sum_{i=1}^{l_{ab}}\frac{exp(\bar{\alpha}_i^a)}{\sum_{k=1}^{l_{ab}}exp(\bar{\alpha}_k^a)}\bm{h}_i,
	\end{split}
\end{equation}
where $\{\omega_d, \bm{W}_d, \bm{U}_d, \bm{M}_d\}$ are trainable parameters. $\bm{g}_t$ is Gaussian distribution centered around $p_t$, $\sigma = \frac{D}{2}$, and  
$ \bm{e}_{l_{ab}} \in \mathbb{R}^{l_{ab}} $ is a row vector of $ 1 $. 
In this operation, we utilize the Gaussian distribution to optimize the original attention value $\bm{\alpha}^a$ so that the model can focus on the important position and its corresponding context, capture the local structure of sentences, and represent the sentence semantic more precisely.

\textbf{Label Prediction }
\label{s:label-prediction}
After finishing the dynamic selections, we first adopt attention pooling to fuse all the selected important parts to generate a locally-aware representation $\bar{\bm{h}}$ from a detailed perspective as follows:
\begin{equation}
	\label{eq:sentence-attention}
	\begin{split}
		\bar{\bm{H}} &= [\bar{\bm{h}}_1, \bar{\bm{h}}_2, ..., \bar{\bm{h}}_{l_{ab}}], \quad
		\bm{\alpha}^b = \bm{\omega}^{\mathrm{T}}\mathrm{tanh}(\bm{W}\bar{\bm{H}} + b), \\
		\bar{\bm{h}} &= \sum_{i=1}^{l_{ab}}\frac{\mathrm{exp}(\alpha_i^b)}{\sum_{k=1}^{l_{ab}} \mathrm{exp}(\alpha_k^b)}\bar{\bm{h}}_i.
	\end{split}
\end{equation} 

\begin{table*}
	\centering
	\caption{Performance (accuracy) of models on different SNLI test sets and SICK test set.}
	\begin{footnotesize}
		\begin{tabular}{lccc} \hline
			\textbf{Model} & \textbf{Full test} & \textbf{Hard test} & \textbf{SICK test}\\ \hline
			(1) CENN~\cite{zhang2017context} & 82.1\% & 60.4\% & 81.8\% \\
			(2) BiLSTM with Inner-Attention~\cite{Liu2016LearningNL} & 84.5\% & 62.7\%  & 85.2\% \\
			(3) Gated-Att BiLSTM~\cite{chen2017recurrent} & 85.5\% & 65.5\% & 85.7\%\\
			(4) CAFE~\cite{Tay2017ACA} & 85.9\% & 66.1\% & 86.1\% \\
			(5) Gumbel TreeLSTM~\cite{choi2018learning} & 86.0\% & 66.7\% & 85.8\% \\
			(6) Distance-based SAN~\cite{im2017distance} & 86.3\% & 67.4\% & 86.7\%\\
			(7) DRCN~\cite{Kim2018SemanticSM} & 86.5\%& 68.3\% & 87.4\% \\ \hline
			(8) DSA~\cite{yoon2018dynamic} & 87.4\% & 71.5\% & 87.7\%  \\ 
			(9) DRr-Net~\cite{zhang2019drr} & 87.5\% & 71.2\% & 87.8\% \\
			(10) BERT-base~\cite{devlin2018bert} & 90.3\% & 80.8\% & \textbf{88.5}\% \\ \hline
			(11) \shortname & \textbf{90.72}\% & \textbf{81.44}\% & 88.36\% \\ 
			\hline
		\end{tabular}
	\end{footnotesize}
	\label{t:snli-result}
\end{table*}

After getting the locally-aware representation $\bar{\bm{h}}$, we leverage heuristic matching~\cite{Chen-Qian2017ACL} between $\bm{h}$ generated from a global aspect and $\bar{\bm{h}}$ generated from a detailed aspect. Then we send the result $\bm{u}$ to a two-layer MLP for final classification. This process is formulated as follows:
\begin{equation}
	\label{eq:matching}
	\begin{split}
		\bm{u} = [\bm{h}_g, \bar{\bm{h}}, \bm{h}_g\odot\bar{\bm{h}}, \bar{\bm{h}}- \bm{h}_g], \quad
		P(y|(\bm{s}^a, \bm{s}^b)) = \mathrm{MLP}(\bm{u}),
	\end{split}
\end{equation}
where concatenation can retain all the information~\cite{zhang2017context}. The element-wise product is a certain measure of ``similarity'' of two sentences~\cite{mou2016natural}. Their difference can capture the degree of distributional inclusion in each dimension~\cite{weeds2014learning}.

\section{Experiment}
\label{s:experiment}
In this section, we first present the details about the model implementation. 
Then, we introduce the datasets that we will evaluate our model on, including four benchmark datasets for two sentence semantic matching tasks, which cover different domains and exhibit different characteristics. 
Next, we will make a detailed analysis about the model and experimental results. 

\subsection{Experimental Setup}
\label{s:model-learning}

\textbf{Loss Function. }
Since sentence semantic matching task can be formulated as classification task, we employ \textit{cross-entropy} as the loss function:
\begin{equation}\label{eq:loss-single}
	\begin{split}
		L = -\frac{1}{N} \sum_{i=1}^{N} \bm{y}_i \mathrm{log} P(y_i | (\bm{s}^a_i, \bm{s}^b_i)) + \epsilon \left\|\bm{\theta}\right\|_2,
	\end{split}
\end{equation}
where $\bm{y}_i$ is the one-hot representation for the true class of the $i^{th}$ instance. $N$ represents the number of training instances. 
$\epsilon$ is the weight decay. 
$\theta$ denotes the trainable parameters in the model and $\left\|\bm{\theta}\right\|_2$ is l2-norm for these parameters.

\begin{table*}
	\begin{floatrow}
		\capbtabbox{
			\begin{footnotesize}
				\begin{tabular}{lcc} \hline
					\textbf{Model} & \textbf{Quora} & \textbf{MSRP} \\ \hline
					(1) CENN~\cite{zhang2017context} & 80.7\% & 76.4\%\\
					(2) MP-CNN~\cite{he2015multi} & - & 78.6\% \\
					(3) BiMPM~\cite{Wang2017BilateralMM} & 88.2\% & -\\
					(4) DRCN~\cite{Kim2018SemanticSM} & 90.2\% & 82.5\%\\ 
					(5) DRr-Net~\cite{zhang2019drr} & 89.7\% & 82.5\% \\
					(6) BERT-base~\cite{devlin2018bert} & 91.1\% & 84.3\%\\ \hline
					(7) \shortname & \textbf{91.7\%} & \textbf{84.5\%} \\ 
					\hline
				\end{tabular}
			\end{footnotesize}
		}{
			\caption{Experimental Results on Quora and MSRP datasets.}
			\label{t:pi-result}
		}
		\capbtabbox{
			\begin{footnotesize}
				\begin{tabular}{lcc} \hline
					\textbf{Model} & \textbf{SNLI test} & \textbf{SICK test} \\ \hline
					(1)~BERT-base & 90.3\% & 88.5\%  \\ 
					(2)~\shortname~(w/o vector $\bm{h}_g$) & 85.3\% & 83.2\%  \\ 
					(3)~\shortname~(w/o vector $\bar{\bm{h}}$) & 89.4\% & 87.5\%  \\ \hline
					(4)~DRr-Net & 87.5\% & 87.8\% \\
					(5)~Multi-GRU + DGA & 88.4\% & 88.1\%  \\ 
					(6)~\shortname~(w/o local context)  & 90.5\% & 88.5\% \\ \hline
					(7)~\shortname~ & \textbf{90.72}\%  & \textbf{88.36}\% \\ 
					\hline
				\end{tabular}
			\end{footnotesize}
		}{
			\caption{Ablation performance (accuracy) of \shortname.}
			\label{t:ablation-result}
		}
		
	\end{floatrow}
\end{table*}

\textbf{Model Initialization. }
We have tuned the hyper-parameters on validation set for best performance.
An \textit{Early-Stop} operation is employed to select the best model. 
Some common hyper-parameters are listed as follows:

The vocabulary is the same as the vocabulary of BERT-base. 
The window size in Gaussian distribution is $D = 4$. 
The dynamic attention length in DGA is $T = 4$.  
The attention size in DGA is set to $200$. 
The hidden state size of GRU is set to $768$. 
The initial learning rate is set to $10^{-4}$. 
An Adam optimizer with $\beta_1 =0.9$ and $\beta_2 = 0.999$ is adopted to optimize all trainable parameters.

\textbf{Dataset.}
In order to evaluate the model performance comprehensively, we employ two sentence semantic matching tasks: \textit{Natural Language Inference~(NLI)} and \textit{Paraphrase Identification~(PI)} to conduct the experiments. 
NLI task requires an agent to predict the semantic relation from premise sentence to hypothesis sentence among ``\textit{Entailment, Contradiction, Neutral}''. 
We select two well-studied and public available datasets: SNLI~\cite{bowman2015large} and SICK~\cite{marelli2014semeval}.
Meanwhile, PI task requires an agent to identify whether two sentences express the same semantic meaning or not. For this task, we select Quora~\cite{iyer2017first} and MSRP~\cite{dolan2005automatically} to evaluate the model performance.

\subsection{Experiment Results}
In this section, we will give a detailed analysis about the models and experimental results. Here, we use \textbf{\textit{Accuracy}} on different test sets to evaluate the model performance.

\textbf{Performance on SNLI and SICK. }
Table~\ref{t:snli-result} reports the results of \shortname~compared with other published baselines. 
We can observe that \shortname~achieves highly comparable performance on different NLI test set. 
Specifically, we make full use of pre-trained language model to get the comprehensive understanding about the semantic meanings. 
This is one of the important reasons that \shortname~is capable of outperforming other BERT-free models by a large margin. 
Furthermore, we develop a novel DGA unit to further improve the capability of dynamic attention mechanism. 
Instead of only selecting one important part at each attention operation, DGA can select the important part and proper local context simultaneously at each step. 
Therefore, the local context of the sentence can be fully explored, and sentence semantics can be represented more precisely. 
This is another reason that \shortname~achieves better performance than all baselines, including the BERT-base model.

Among all baselines, DRr-Net~\cite{zhang2019drr} and DSA~\cite{yoon2018dynamic} are current state-of-the-art methods without BERT. 
DSA~\cite{yoon2018dynamic} modifies the dynamic routing in capsule network and develops a DSA to model sentences. 
It utilizes CNN to capture the local context information and encodes each word into a meaningful representation space. 
DRr-Net adopts multi-layer GRU to encode the sentence semantic meanings from a global perspective and designs a dynamic re-read attention to select one important part at each attention step for detailed sentence semantic modeling. 
They all achieved impressive performances. 
However, both RNN and CNN structures have some weaknesses in extracting features or generating semantic representations compared with BERT. 
We can observe from Table~\ref{t:snli-result} that the BERT-base model outperforms them by a large margin. 
Meanwhile, their attention operations either select too many important parts at one time or only focus on one important part at each operation, which may lead to a misunderstanding of the sentence semantic meanings. 
Thus, their performance is not as good as \shortname~reaches. 
On the other hand, apart from the powerful encoding ability, BERT still focuses on the importance of words to the sequence and has some weaknesses in distinguishing the exact meanings of sentences. 
By taking the local context into consideration and leveraging DGA to get the precise meanings of sentences, \shortname~is able  to achieve better performance than BERT.

\textbf{Performance on Quora and MSRP. }
Besides NLI task, we also select PI task to better evaluate the model performance on sentence semantic similarity identification.
Table~\ref{t:pi-result} illustrates the experimental results on Quora and MSRP datasets. 
Different from the results on NLI datasets, our proposed \shortname~achieves the best performance compared with other baselines on both test sets, revealing the superiority of our proposed \shortname. 
Besides, we can obtain that almost all the methods have better performance on Quora dataset and the improvement of our proposed \shortname~on Quora dataset is also larger than the improvement on MSRP dataset. 
Quora dataset~\cite{iyer2017first} has more than 400k sentence pairs, which is much larger than MSRP dataset. 
Large data is capable of helping to model to better analyze the data and get close to the upper bound of the performance.
Meanwhile, we also speculate that the inter-sentence interactions is probably another possible reason. 
Quora dataset contains many sentence pairs with less complicated interactions~(e.g., many identical words in two sentences)~\cite{lan2018neural}.

\subsection{Ablation Performance}
The overall performance has proven the superiority of \shortname. However, which part is more important for performance improvement is still unclear. 
Thus, we conduct an ablation study on two NLI test sets to examine the effectiveness of each component. 
Recall the model structure, two important semantic representations are $\bm{h}_g$ from BERT output and $\bar{\bm{h}}$ from DGA output. 
As illustrated in Table~\ref{t:ablation-result}(2)-(3), when we remove the global representation $\bm{h}$, we can observe that the model performance has a big drop. This result is in line with our intuitive. We should have a comprehensive understanding about the sentence before making a decision. Only the important parts are insufficient for the decision making. 

Meanwhile, when removing the detailed representation $\bar{\bm{h}}$, model performance is worse than BERT-base model. we speculate that DGA is in the training process but not in the predicting process, which decreases the model performance. 
Besides, we investigate the effectiveness of BERT encoder and local context. When replacing BERT with multi-layer GRUs, we can observe that its performance is still better than DRr-Net, suggesting the importance of local context utilization. Meanwhile, its performance is not comparable with BERT-base, let alone the entire \shortname, proving the importance of BERT. 
When removing the local context, the performance of \shortname~is capable of optimizing the BERT-base model, proving the effectiveness of local context utilization. 
In other words, both BERT encoder and local context utilization are indispensable for \shortname~to achieve better performance. 

\begin{figure*}
	\centering
	\includegraphics[width=0.90\textwidth]{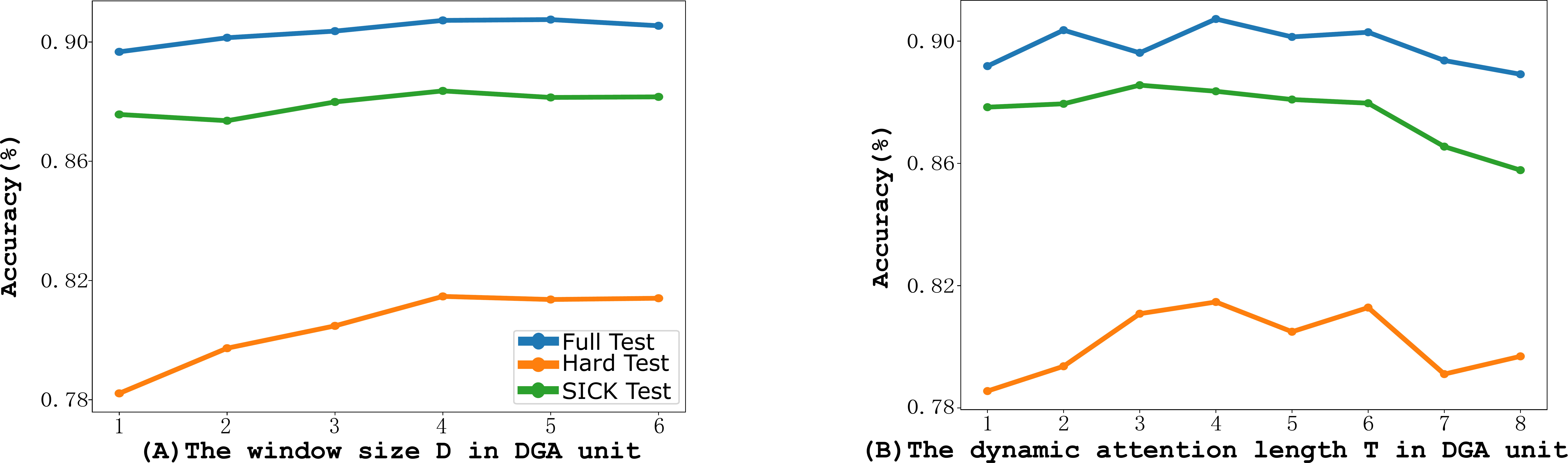}
	\caption{Performance of \shortname~with different window sizes~(1-6), and attention lengths~(1-8).}
	\label{f:sensitive}
\end{figure*}

\subsection{Sensitivity of Parameters}
There are two hyper-parameters that affect the model performance: 1) The window size $D$ in DGA unit; 2) The dynamic attention length $T$ in DGA unit. Therefore, we evaluate \shortname~performance on two NLI test sets with different hyper-parameter settings. The results are summarized in Figure~\ref{f:sensitive}.

When talking about the window size in DGA unit, we can observe that the model performance first increases and then becomes smooth with the increase of window size. 
We speculate that a too small or too big window cannot help to capture the local structure for precisely semantic understanding. 
When the window size is $D = 4$, DGA will consider two words on each side of the center word, which is 
suitable for leveraging local context to enhance the semantic understanding of sentences.

As for the dynamic attention length, Bowman et al.~\cite{bowman2015large} has conducted that the average length is $14.1$ for premise and $8.3$ for hypothesis in SNLI. 
From Figure~\ref{f:sensitive}(B), $4$ is suitable for dynamic attention length. 
Too short reading length may cause the model to ignore some important parts. Meanwhile, too long reading length may weaken the ability of precisely local structure capturing and semantic understanding.

\section{Conclusion and Future Work}
\label{s:conclusion}
In this paper, we proposed an effective \fullname~approach for sentence semantic matching, a novel architecture that not only models sentence semantics in a global perspective, but also utilizes local structure to support the analysis of the important parts step by step. 
To be specific, we first make full use of pre-trained language model to evaluate semantic meanings of words and sentences from a global perspective. 
Then, we design a novel Dynamic Gaussian Attention~(DGA) to pay close attention to one important part and corresponding local context among sentences simultaneously at each attention operation. 
By taking the local information into consideration, \shortname~is capable of measuring the sentence semantics more comprehensively. 
Finally, we integrate the global semantic representation from Bert and detailed semantic representation from DGA to further improve the model performance on sentence semantic matching. 
Extensive evaluations on two sentence semantic matching tasks (i.e., NLI and PI) demonstrate the superiority of our proposed \shortname.  
In the future, we will focus on providing more information for dynamic attention to better local important parts selecting and sentence semantic understanding.

\section*{Acknowledgements}
This work was supported in part by grants from the National Natural Science Foundation of China (Grant No. 62006066), the Open Project Program of the National Laboratory of Pattern Recognition (NLPR), and the Fundamental Research Funds for the Central Universities, HFUT.

\bibliographystyle{splncs04}
\bibliography{7_myReference}

\end{document}